\documentclass{article}
\usepackage{graphicx} 
\usepackage{pgfplots}
\pgfplotsset{compat=1.18}
\usepackage{outlines}
\usepackage{multicol}
\usepackage{enumitem}
\usepackage{hyperref}
\usepackage{xcolor}
\usepackage{amsmath}
\usepackage{natbib}
\usepackage{booktabs}
\usepackage{amssymb}
\usepackage{pifont}
\newcommand{\cmark}{\ding{51}} 
\newcommand{\xmark}{\ding{55}} 

\usepackage[preprint]{neurips_2025}
\usepackage[utf8]{inputenc} 
\usepackage[T1]{fontenc}    
\usepackage{hyperref}       
\usepackage{url}            
\usepackage{booktabs}       
\usepackage{amsfonts}       
\usepackage{nicefrac}       
\usepackage{microtype}      
\usepackage{xcolor}         

\title{W4S4: WaLRUS Meets S4 for Long-Range Sequence Modeling}

\author{%
  Hossein Babaei\\
  Department of Electrical and Computer Engineering\\
  Rice University\\
  \texttt{hb26@rice.edu} 
   \And
   Mel White \\
   Department of Electrical and Computer Engineering\\
   Rice University\\
   \texttt{mel.white@rice.edu} 
   \And
   Richard G. Baraniuk \\
   Department of Electrical and Computer Engineering\\
   Rice University\\
   \texttt{richb@rice.edu} \\
   }
   
\begin{document}

\maketitle

\begin{abstract}
State Space Models (SSMs) have emerged as powerful components for sequence modeling, enabling efficient handling of long-range dependencies via linear recurrence and convolutional computation. However, their effectiveness depends heavily on the choice and initialization of the state matrix. In this work, we build on the SaFARi framework and existing WaLRUS SSMs to introduce a new variant, \textbf{W4S4} (WaLRUS for S4), a new class of SSMs constructed from redundant wavelet frames. WaLRUS admits a stable diagonalization and supports fast kernel computation without requiring low-rank approximations, making it both theoretically grounded and computationally efficient. We show that WaLRUS retains information over long horizons significantly better than HiPPO-based SSMs, both in isolation and when integrated into deep architectures such as S4. Our experiments demonstrate consistent improvements across delay reconstruction tasks, classification benchmarks, and long-range sequence modeling, confirming that high-quality, structured initialization enabled by wavelet-based state dynamic offers substantial advantages over existing alternatives. WaLRUS provides a scalable and versatile foundation for the next generation of deep SSM-based models.
\end{abstract}

\section{Introduction}

Time series and sequence modeling are central to a broad range of applications, including speech recognition, medical diagnostics, and financial forecasting. Over the years, a diverse set of architectures have been developed to tackle these tasks, including linear autoregressive models, multilayer perceptrons (MLPs), convolutional neural networks (CNNs), recurrent neural networks (RNNs), and more recently, Transformers. \cite{wen2022transformers,alemohammad2019wearing,masini2023machine,liu2022scinet}

Each of these methods, however, comes with distinct limitations. Autoregressive models are interpretable and easy to train but lack the capacity to model complex dependencies in real-world data \cite{zhang2003time}. MLPs are highly expressive but do not encode temporal structure, and their dense parameterization often leads to overfitting, particularly for long sequences.\cite{chen2023tsmixer}

CNNs and RNNs offer classical solutions that mitigate these issues by introducing temporal inductive biases. RNNs extend MLPs with hidden states that evolve over time, enabling temporal modeling. Yet, they suffer from poor parallelization \cite{van2019rethinking} and are prone to vanishing or exploding gradients, which limit their ability to capture long-range dependencies\cite{pascanu2013difficulty}. CNNs, in contrast, are linear, inherently parallelizable, and stable during training. However, their receptive field grows slowly with depth \cite{luo2016understanding}, and convolution operations remain fundamentally linear unless supplemented with non-linearities, which can limit expressiveness.

State Space Models (SSMs) have recently re-emerged as a compelling alternative for long-range sequence modeling. SSMs apply linear recurrence relations to a hidden state, blending favorable properties from both RNNs and CNNs. Like RNNs, they support recursive processing and streaming inference with low memory and latency. Like CNNs, they allow fast convolution through kernel fusion and are parallelizable on modern hardware. Crucially, SSMs parameterize long convolutional kernels compactly, enabling them to summarize long histories efficiently without overfitting.

While Transformer models have achieved remarkable success in natural language processing and have been adapted to time-series domains, they also have notable drawbacks. Self-attention mechanisms scale quadratically with input length, incurring high computational and memory costs. Moreover, Transformers are inherently permutation-invariant and must rely on external positional embeddings to encode time, which introduces a structure that is not native to the model.

These limitations are exacerbated in high-sampling-rate signals, where adjacent samples are highly redundant. In such contexts, Transformers suffer from inefficiency and a lack of structured tokenization strategies, an open challenge in the field.

In contrast, SSMs naturally encode temporal order and summarize long signal histories with high fidelity. However, their expressiveness is constrained by the linear nature of their recurrence. Their parameter efficiency stems from structural constraints on the kernel, which may limit flexibility compared to CNNs with unconstrained long filters.

The effectiveness of SSMs ultimately depends on the quality of their state representations. Previous studies observed that SSMs with randomly initialized parameters often fail to outperform standard CNNs or RNNs. This performance gap is rooted in the fact that downstream nonlinear layers in SSM-based networks operate only on the hidden state; thus, the quality of this representation is paramount.

The HiPPO framework \cite{gu2020hippo} established a connection between online function approximation and SSMs by projecting signals onto polynomial bases, specifically Legendre polynomials. This allowed the SSM to maintain a compressed yet rich representation of signal history. Notably, even when the HiPPO matrices were fixed (i.e., not trained), the resulting network achieved state-of-the-art performance on long-range benchmarks, underscoring the importance of initialization.

The S4 architecture \cite{gu2022s4} generalized HiPPO-based SSMs by proposing structured and diagonalizable variants, improving training stability and scalability. Subsequent work found that the choice of initialization significantly impacts performance across datasets.

Building on this, the SaFARi framework \cite{babaei2025safari} extended SSM initialization to arbitrary algebraic frames, allowing more flexible and expressive state encodings. A notable example is WaLRUS \cite{babaei2025walrus}, which instantiated SaFARi using wavelet frames to achieve improved compactness and fidelity.

\paragraph{Our Contributions.}

In this paper
\begin{itemize}
    \item we build on SaFARi and S4 to propose a new initialization scheme for state space models, which we call \textbf{W4S4 (WaLRUS for S4)}. Our approach leverages the representational advantages of redundant wavelet frames to enhance the quality of the state vector used in SSMs.
    \item We evaluate W4S4 on benchmark tasks and datasets introduced in the S4 paper. Our experiments demonstrate that W4S4 consistently outperforms S4 models initialized with HiPPO, even when the state matrices ($A$, $B$) are not trained. We attribute this performance gain to improved state representation, which alleviates the representational bottleneck inherent in SSM-based architectures.

\end{itemize}

The remainder of this paper is organized as follows. Section~\ref{sec:background} reviews foundational models for sequence processing, including RNNs, CNNs, Transformers, and State Space Models, and introduces the HiPPO and SaFARi frameworks that motivate our work. In Section~\ref{sec:method}, we present the construction of WaLRUS from redundant wavelet frames and explain its integration into deep SSM architectures. Section~\ref{sec:ablation} provides controlled ablation studies comparing WaLRUS to HiPPO in terms of memory retention and delay reconstruction. Section~\ref{sec:experiments} evaluates the performance of WaLRUS in full deep learning pipelines across several benchmarks. We conclude in Section~\ref{sec:conclusion} with a discussion of implications and future directions.

\section{Background} \label{sec:background}

This section reviews key developments in sequence modeling, focusing on the progression from RNNs and CNNs to State Space Models (SSMs) and Transformer-based architectures. We highlight the limitations of traditional models, introduce the HiPPO framework for online function approximation, and discuss its generalization via the SaFARi framework. We then examine the critical role of initialization in SSMs and describe how these ideas culminate in the S4 model, a scalable and expressive architecture for long-range sequence tasks.

\subsection{Recurrent and Convolutional Neural Networks}
Recurrent Neural Networks (RNNs) model sequential data by maintaining a hidden state updated at each time step via $h_t = f(h_{t-1}, x_t)$. This enables constant-memory processing and temporal dependency modeling. However, RNNs struggle with vanishing and exploding gradients during backpropagation through long sequences, hindering their ability to capture long-range dependencies~\cite{patro2024mamba}. Gated variants such as LSTMs and GRUs mitigate these issues but remain difficult to train over extended contexts. Moreover, RNNs are inherently sequential, limiting parallelism and making them inefficient on modern hardware.

Convolutional Neural Networks (CNNs) apply local filters across the sequence, enabling parallel computation and fast training. While effective at capturing short-term patterns, CNNs lack persistent state and must stack multiple layers or use dilation to reach long-range context. This leads to inefficiency in online or streaming settings, where overlapping windows require recomputation or buffering. In essence, RNNs offer dynamic memory but poor scalability, while CNNs are scalable but stateless. This contrast motivates the search for architectures that combine the strengths of both.

\subsection{State Space Models: Linear Recurrences and Efficient Inference}

State Space Models (SSMs) provide a principled framework for sequence modeling through linear dynamical systems. The system evolves via a hidden state $x(t)$ governed by:

\begin{equation*}
    \dot{x}(t) = -A x(t) + B u(t)
\end{equation*}
\begin{equation*}
    y(t) = C x(t) + D u(t),
\end{equation*}

where $u(t)$ is the input, $x(t)\in\mathbb{R}^N$ is the state, and $y(t)$ is the output. Discretization yields the update rule $x_{n+1} = A_d x_n + B_d u_n$, with $A_d \approx e^{A\Delta t}$, forming a linear recurrent model. Crucially, stability is ensured by choosing $A$ with negative real parts or $A_d$ with eigenvalues inside the unit circle, allowing the model to handle long sequences without gradient instability—an advantage over nonlinear RNNs.

SSMs can be viewed in both recurrent and convolutional forms. At inference time, they can operate recurrently with constant-time updates per step, enabling stateful, low-latency streaming. Alternatively, due to linearity and time-invariance, SSMs define a convolution kernel $k(n)$ such that $y_n = (u * k)_n$, allowing parallel training using FFT-based or fused-kernel convolution~\cite{patro2024mamba}. This duality enables efficient training with $O(L)$ complexity, unlike attention-based methods with $O(L^2)$ cost. The convolutional view accelerates training, while the recurrent view supports deployment in real-time systems.

Another strength of SSMs is their adaptability to continuous time and irregular sampling. Since the model arises from a continuous-time ODE, varying the step size $\Delta t$ requires no retraining, making SSMs ideal for real-world data with nonuniform sampling~\cite{gu2021combining}.

In summary, SSMs offer an efficient and stable alternative for long-range sequence modeling by combining the recurrent benefits of memory and low latency with the parallelism and scalability of convolution.

\subsection{Transformer Models and Their Limitations}

Transformers revolutionized sequence modeling by replacing recurrence with self-attention, allowing each token to directly attend to all others. This enables efficient modeling of long-range dependencies without step-by-step propagation, and the fully parallelizable attention mechanism dramatically accelerates training compared to RNNs. These advantages led to widespread adoption across NLP and beyond, following the success of the "Attention Is All You Need" architecture~\cite{vaswani2017attention}.

Despite their success, Transformers face key limitations with long sequences and online tasks. The self-attention mechanism incurs $O(L^2)$ time and memory complexity due to pairwise interactions between all $L$ tokens, making it impractical for very long inputs. Even efficient variants struggle to scale beyond a few thousand tokens. Moreover, Transformers are inherently position-agnostic and rely on external positional encodings (fixed or learned) to model token order. This adds design complexity and limits generalization to longer unseen sequences. Lastly, Transformers require full-sequence access, making them ill-suited for streaming or online scenarios. Unlike RNNs or SSMs, they cannot natively process inputs incrementally without costly workarounds. These constraints have prompted renewed interest in SSM-based models, which offer a more scalable and stateful alternative for handling long-range dependencies.

\subsection{The HiPPO Framework for Online Function Approximation}

The HiPPO (High-Order Polynomial Projection Operators) framework~\cite{gu2020hippo} addresses the problem of maintaining a compressed, online representation of the input history. It formulates this as a function approximation task: at each time $t$, the goal is to project the past signal onto a polynomial basis while minimizing approximation error, typically with a time-dependent weighting that prioritizes recent inputs.

HiPPO achieves this by projecting the input onto an orthogonal polynomial basis—most commonly the Legendre polynomials. As new data arrives, the coefficients representing the past are updated via a linear differential equation:
\[
\dot{x}(t) = A_{\text{HiPPO}} x(t) + B_{\text{HiPPO}} u(t),
\]
where $x(t)$ are the basis coefficients. The choice of the host basis (e.g. Legendre polynomial) and weighting measure $w(t)$ (e.g., exponential decay) determines the specific $A, B$ matrices. This yields a principled SSM that optimally compresses past inputs. The Legendre Memory Unit (LMU) was shown to be a special case, with HiPPO-LegS providing a scale-invariant memory that compresses the entire history into polynomial coefficients over $[0, t]$.

HiPPO-LegS has several desirable properties: continuous-time formulation, stability, timescale robustness, and efficient linear updates. It avoids vanishing or exploding gradients and generalizes recurrence by analytically deriving updates from approximation theory. These properties made HiPPO a foundational tool in structured state-space models like S4, where the HiPPO matrix is used to initialize or define the recurrent dynamics.

In summary, HiPPO provides a mathematically grounded design for SSMs with optimal memory retention. By replacing heuristic or randomly initialized recurrence with structured dynamics derived from polynomial projection theory, HiPPO has inspired a new class of efficient and memory-capable sequence models.

\subsection{Beyond HiPPO: The SaFARi Framework and the WaLRUS SSMs}

While HiPPO introduced a powerful paradigm for online function approximation through orthogonal polynomial projections, it remains limited to a narrow class of bases, most notably the Legendre polynomials. To expand the design space of SSMs, the SaFARi framework~\cite{babaei2025safari} generalizes the construction of HiPPO by allowing state-space dynamics to be derived from arbitrary algebraic frames, not just orthogonal bases. SaFARi formalizes the connection between frame-theoretic signal approximation and continuous-time state-space systems, enabling the creation of infinitely many “species” of SSMs, each inheriting distinct inductive biases and spectral properties from their underlying frame.

One such species is \textbf{WaLRUS}~\cite{babaei2025walrus}, a SaFARi-based SSM constructed using redundant wavelet frames. Unlike HiPPO-LegS, which lacks a stable diagonalization, WaLRUS admits an exact and well-conditioned diagonal form. This property allows for fast kernel computation without requiring Diagonal-Plus-Low-Rank (DPLR) approximations or heuristic optimization schemes. Moreover, wavelet frames offer a natural multiresolution decomposition of signals, endowing WaLRUS with strong localization in both time and frequency. As we demonstrate later in this work, these characteristics translate into superior memory retention and downstream task performance, positioning WaLRUS as a practical and theoretically grounded alternative to traditional HiPPO-based SSMs.

\subsection{Impact of Parameter Initialization in SSM-Based Models}

The performance of SSM-based sequence models critically depends not only on model design but also on initialization, particularly of the state matrix $A$ which governs memory dynamics. Poorly initialized $A$ (e.g., random or overly decaying) often leads to unstable training or weak long-term memory. Empirical studies~\cite{gu2021combining} showed that random diagonal or dense $A$ matrices perform significantly worse on long-range tasks compared to structured matrices like HiPPO, motivating their adoption in models like S4.

The HiPPO-initialized $A$ matrix endows the SSM with a structured set of basis functions (e.g., Legendre polynomials), enabling it to encode inputs over a broad range of timescales from the start. This allows the SSM to act as a strong memory module even before training, easing optimization and improving generalization. In contrast, random initializations may lack critical temporal modes or lead to rapid forgetting, impairing the model’s ability to learn long dependencies.

Recent work~\cite{gu2022parameterization} demonstrated that diagonal SSMs (e.g., S4D) can match full S4 performance if their eigenvalues are initialized to approximate the HiPPO spectrum. This initialization, termed HiPPO-LegS-D, ensures that the induced convolution kernel mirrors that of S4 in the large state limit. These findings reinforce the idea that initialization is not just a training aid, but a functional prior: it supplies a useful set of temporal filters that gradient descent can refine. Without it, the model may never discover essential dynamics.

In summary, initialization of $A$ is central to SSM model success. Structured choices like the HiPPO matrix or its diagonalized variants provide stability, long-range memory, and inductive biases that are difficult to recover through training alone. Modern SSMs thus prioritize both expressive parameterizations and principled initialization strategies to ensure robust performance across tasks.

\subsection{The S4 Model Architecture for Long-Range Sequence Modeling}

The Structured State Space Sequence Model (S4) integrates state-space modeling into a deep neural architecture analogous to a Transformer. Each S4 layer replaces the attention mechanism with a linear state-space layer (SSM), followed by a standard feed-forward network (FFN), residual connections, and normalization. This structure enables S4 to efficiently model long sequences using the SSM's convolutional form, while the FFN adds nonlinearity and expressivity.

Formally, each layer computes $y = \text{SSM}_{\theta}(u)$ via convolution using a structured state matrix $A$, typically initialized with a HiPPO-derived form. The output is passed through a position-wise MLP, with residual and normalization layers applied around both components. This design, supported by Koopman operator theory, allows linear SSMs to capture global dynamics, while nonlinear FFNs provide approximation power for complex sequence behaviors.

A key advantage of S4 is that it does not require explicit positional encodings. The SSM inherently respects temporal order via its recurrence or convolution, making the model sequence-aware by design. Moreover, S4 achieves high efficiency by using a diagonal-plus-low-rank parameterization of $A$, enabling fast kernel computation through analytical methods (e.g., via Cauchy kernel techniques). This results in $O(L \log L)$ time complexity and significantly faster autoregressive inference than Transformers—up to 60$\times$ speedup in some settings.

S4 achieves strong performance across long-range tasks. On the Long Range Arena (LRA) benchmark, it matched or exceeded Transformer performance, including solving the challenging 16k-length Path-X task. It also performed competitively on sequential CIFAR-10, achieving ResNet-level accuracy with a purely sequential architecture. Ablations confirm that the HiPPO-based initialization of $A$ is crucial, as replacing it with random matrices leads to substantial performance degradation.

In summary, S4 retains the deep residual architecture of Transformers while replacing attention with a structured and scalable state-space layer. Its HiPPO-based design enables long-range memory, and its convolutional implementation allows efficient parallel training and fast inference. As subsequent models like S4D and S5 build on these principles, SSM-based architectures continue to emerge as strong alternatives for long-sequence modeling.

\section{method} \label{sec:method}
We propose \text{W4S4}, a novel State Space Model (SSM) that serves as a drop-in replacement for HiPPO-based initializations in deep sequence models such as S4, S5, and other architectures with SSM cores. WaLRUS is constructed using the SaFARi framework~\cite{babaei2025safari}, but instead of polynomial bases, it leverages redundant wavelet frames~\cite{babaei2025walrus}.

Unlike HiPPO-LegS and HiPPO-LegT, WaLRUS admits a stable diagonal representation, avoiding the need for Diagonal-Plus-Low-Rank (DPLR) approximations and the associated computational costs for kernel computation. This makes WaLRUS more efficient in both memory and runtime, especially in deep architectures.

This section outlines the construction of WaLRUS, its integration into existing deep SSM structures, and provides a detailed analysis of its computational and memory complexity compared to HiPPO variants.

\subsection{W4S4: Scaled-WaLRUS matrices in time invariant form}

We define \textbf{W4S4}—short for \emph{WaLRUS for S4} and phonetically aligned with the original name—as the variant of WaLRUS that is used for S4 initialization. Depending on the chosen measure (scaled or translated), the corresponding WaLRUS SSM takes the following form:

\begin{equation}
    \frac{d}{dt} x(t) = -\frac{1}{t} A_{\mathrm{sc}} x(t) + \frac{1}{t} B_{\mathrm{sc}} u(t),
\end{equation}
\begin{equation}
    \frac{d}{dt} x(t) = -\frac{1}{\theta} A_{\mathrm{tr}} x(t) + \frac{1}{\theta} B_{\mathrm{tr}} u(t).
\end{equation}

Following \cite{gu2022train}, it is common to construct a time-invariant SSM by reusing $A_{\mathrm{sc}}$ and $B_{\mathrm{sc}}$ in a translated formulation:

\begin{equation}
    \frac{d}{dt} x(t) = -\frac{1}{\theta} A_{\mathrm{sc}} x(t) + \frac{1}{\theta} B_{\mathrm{sc}} u(t).
    \label{eq:tissm}
\end{equation}

This formulation effectively defines an SSM with basis functions given by exponentially warped Legendre polynomials, i.e., $p_n(t) = L_n(e^{-t})$. The translated dynamic yields two practical advantages:
\begin{itemize}
    \item The differential equation becomes time-invariant, allowing the convolution kernel to be precomputed and reused across different sequence lengths.
    \item The exponential warping gives the basis functions infinite support while prioritizing recent history, which is a desirable property for modeling long-range dependencies.

\end{itemize}

Empirically, all HiPPO-based deep SSMs adopt the translated form for computational efficiency. Hence, S4-LegS and S4-LegT share the same structural formulation, differing only in their $A, B$ matrices.

We follow this paradigm to define W4S4: we first construct a WaLRUS SSM using the scaled measure, then adopt the time-invariant form in Eq.~\ref{eq:tissm}. This construction allows W4S4 to be seamlessly integrated into existing SSM-based architectures, while inheriting the stability and efficiency of wavelet-based diagonalization.

\subsection{Redundant Frames and Dimensionality Reduction}

WaLRUS is distinguished from prior HiPPO models by its use of redundant wavelet frames instead of orthogonal polynomial bases. This introduces a higher-dimensional state space, raising concerns about increased memory and computational costs.

However, WaLRUS benefits from a key structural property that mitigates these concerns.

\paragraph{Theorem 1.} \textit{(See Appendix for proof.)}
Let $A$ be the diagonalized state matrix of WaLRUS constructed from a redundant wavelet frame. Then:
\begin{itemize}
    \item Only the first $N_{\mathrm{eff}}$ eigenvalues of $A$ are greater than 1; the rest are exactly 1.
    \item Only the first $N_{\mathrm{eff}}$ elements of the state vector $\widetilde{x}$ (in the diagonal basis) contribute to the reconstruction of the input signal.
    \item The kernel $K$ can be computed using only the top $N_{\mathrm{eff}}$ rows of $\widetilde{K}$ and the corresponding rows of the eigenvector matrix $V$.
\end{itemize}

This result allows us to truncate the WaLRUS state space to $N_{\mathrm{eff}}$ dimensions without sacrificing reconstruction fidelity. In practice, we:
\begin{enumerate}
    \item Diagonalize $A$ and compute only the top $N_{\mathrm{eff}}$ eigenvalues and rows of $\widetilde{B}$.
    \item Compute the kernel $\widetilde{K}$ using this reduced system.
    \item Absorb $V[0{:}N_{\mathrm{eff}}, :]$ into the output matrix $C$ and proceed with standard SSM computations.
\end{enumerate}

\paragraph{Complex-valued Efficiency.}
Although diagonalized SSMs result in complex-valued kernels, the real-valued nature of $A$ ensures that all eigenvalues appear in complex-conjugate pairs. This symmetry allows us to store only half of the complex values, effectively representing $N_{\mathrm{eff}}$ real dimensions with $\frac{N_{\mathrm{eff}}}{2}$ complex numbers. Thus, the use of complex arithmetic does not increase the memory footprint.

In summary, WaLRUS achieves efficient representation and computation by leveraging the redundancy in wavelet frames and exploiting the spectral structure of its state matrix. This makes it an effective and practical choice for deep learning models with SSM cores.

\section{Ablation Studies: HiPPO vs WaLRUS SSM cores, who remembers better?} \label{sec:ablation}

As discussed in the background section, the primary bottleneck in deep SSM architectures such as S4 lies in the convolutional SSM module. This component is solely responsible for modeling temporal dependencies, effectively functioning as an online memory that provides long-range context.

To isolate and evaluate the memory retention ability of the SSM core, we compare HiPPO and WaLRUS in a controlled delay-reconstruction task. Specifically, we construct a minimal linear model containing only $(A, B, C, D)$ components, with $(A, B)$ fixed to either HiPPO or WaLRUS, and optimize $(C, D)$ to reconstruct a delayed version of the input signal.

Given a fixed delay of $L_0$ samples, we minimize the error between the network output and the input delayed by $L_0$. This yields a convex optimization problem in $(C, D)$, allowing us to determine the optimal loss achievable for each $(A, B)$. This minimal loss directly reflects the memory retention capability of the SSM.

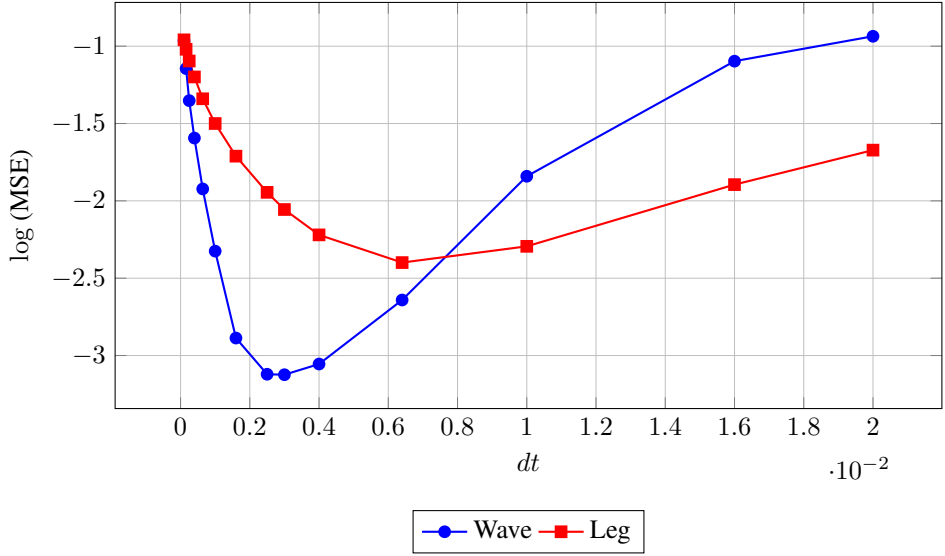
\begin{figure}[t]
    \centering
    \begin{tikzpicture}
    \begin{axis}[
        width=0.9\textwidth,
        height=0.5\textwidth,
        xlabel={$ dt $},
        ylabel={ $\log$ (MSE)},
        legend style={at={(0.5,-0.25)}, anchor=north, legend columns=2},
        grid=both,
        grid style={line width=.1pt, draw=gray!30},
        major grid style={line width=.2pt, draw=gray!50},
    ]
    \addplot[
        color=blue,
        thick,
        mark=*,
    ] table [
        col sep=comma,
        x=dt,
        y expr={-\thisrow{wave}}
    ] {CSV_files/exp_ablation_1.csv};
    \addlegendentry{Wave}

    \addplot[
        color=red,
        thick,
        mark=square*,
    ] table [
        col sep=comma,
        x=dt,
        y expr={-\thisrow{leg}}
    ] {CSV_files/exp_ablation_1.csv};
    \addlegendentry{Leg}
    \end{axis}
\end{tikzpicture}
    \caption{Log-MSE comparison between Scaled-WaLRUS and HiPPO-LegS for various values of $dt = 1/\theta$ when the network parameters are optimized. Scaled-WaLRUS achieves a $0.725$ smaller log-MSE, outperforming HiPPO-LegS by a factor of $\times$\textbf{5.31}.}
    \label{fig:ablation_1}
\end{figure}

Following standard practice, we restrict our analysis to the time-invariant formulation (TOSSM), which includes a tunable hyperparameter $\theta$. In our first study, we fix $L_0 = 300$ and sweep over $\theta \in [10^{-4}, 10^{-2}]$, measuring the optimal log-MSE achieved by each SSM. Results are shown in Fig.~\ref{fig:ablation_1}.

The results show that WaLRUS achieves a lower MSE of $7.5 \times 10^{-4}$ compared to $4 \times 10^{-3}$ for HiPPO, confirming WaLRUS's superior memory retention in the delay task.

In a second study, we vary the delay amount $L_0$ and compare both methods across different delay horizons. For each setting, we select the optimal values for $\theta$, $C$, and $D$ individually.

\begin{figure}[t]
    \centering
    \begin{tikzpicture}
    \begin{axis}[
        width=0.9\textwidth,
        height=0.5\textwidth,
        xlabel={Delay amount},
        ylabel={$\log$(MSE)},
        legend style={at={(0.5,-0.25)}, anchor=north, legend columns=2},
        grid=both,
        grid style={line width=.1pt, draw=gray!30},
        major grid style={line width=.2pt, draw=gray!50},
    ]
    \addplot[
        color=blue,
        thick,
        mark=*,
    ] table [
        col sep=comma,
        x=delay,
        y expr={-\thisrow{wave}}
    ] {CSV_files/exp_ablation_2.csv};
    \addlegendentry{Wave}

    \addplot[
        color=red,
        thick,
        mark=square*,
    ] table [
        col sep=comma,
        x=delay,
        y expr={-\thisrow{leg}} 
    ] {CSV_files/exp_ablation_2.csv};
    \addlegendentry{Leg}
    \end{axis}
\end{tikzpicture}
    \caption{Log-MSE across different delay lengths with optimal $\theta$, $C$, and $D$. Scaled-WaLRUS consistently outperforms HiPPO-LegS in capturing delayed input information.}
    \label{fig:ablation_2}
\end{figure}
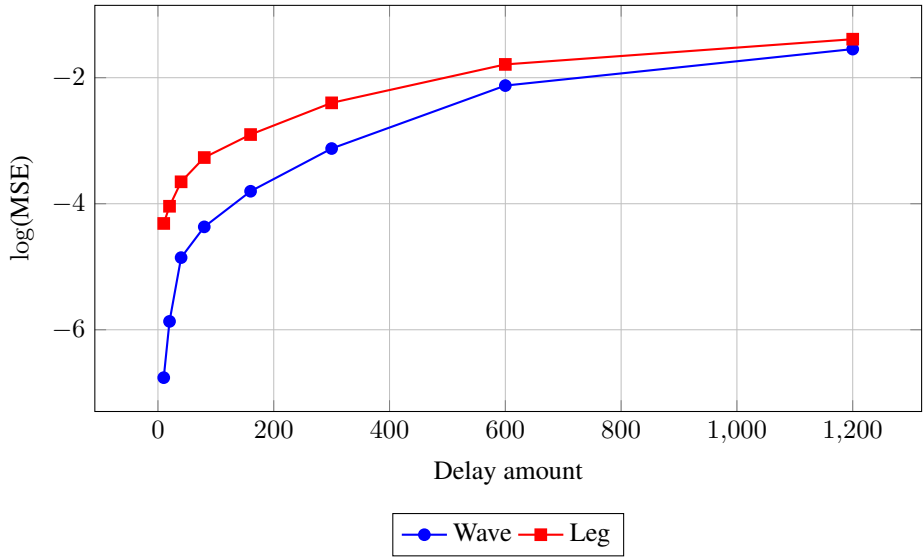

Across both experiments, the results consistently demonstrate the superior memory capabilities of WaLRUS, supporting its use as a high-quality initialization in deep SSM models like S4.

\section{Experiments} \label{sec:experiments}

\subsection{Datasets}

We evaluate WaLRUS across a diverse range of benchmarks, covering vision, audio, biomedical, and time series domains. These include:
\begin{itemize}
    \item Long-Range Arena (LRA) \cite{tay2020long}
    \item Speech Commands \cite{warden2018speech}
    \item Sequential CIFAR \cite{krizhevsky2009learning}
    \item BIDMC Vital Signs \cite{goldberger2000physiobank, pimentel2016toward}
    \item Multivariate forecasting datasets: Electricity, Traffic, Weather, and Solar-Energy \cite{wu2021autoformer, lai2018modeling}
\end{itemize}
Full dataset descriptions and task specifications are included in Appendix~\ref{appendix:datasets}.

\subsection{HiPPO vs WaLRUS in S4: Delay Task}

We now examine the memory performance of WaLRUS and HiPPO when embedded in full S4 models. Delay tasks are considered too simple for deep S4 structures, as both models perform well. To increase task sensitivity, we restrict our S4 architecture to a single layer ($L=1$) with only $H=4$ SSM cores.

All parameters in the S4 model are trainable except $A$, which is initialized and frozen to either HiPPO or WaLRUS. Following \cite{gu2022parameterization}, we also fix $B$ as an all-ones vector, given that the SSM output depends only on the Hadamard product $B \circ C$. Therefore, $B$ and $C$ can effectively be fused together and be learned as $C$.

Table~\ref{tab:delay} reports log-MSE values for various datasets. For the Texts dataset, we report cross-entropy instead.

\begin{table}[ht]
\centering
\begin{tabular}{lccc}
\hline
\textbf{Dataset} & \textbf{HiPPO} & \textbf{WaLRUS} & \textbf{Relative} \\
\hline
M4         & -2.5026 & \textbf{-2.8925} & $\times$ 2.45 \\
Speech     & -0.6566 & \textbf{-0.7761} & $\times$ 1.32 \\
s-CIFAR    & -1.1750 & \textbf{-1.3510} & $\times$ 1.50 \\
Solar      & -1.9450 & \textbf{-2.0070} & $\times$ 1.15 \\
BIDMC      & -0.9220 & \textbf{-0.9420} & $\times$ 1.05 \\
\midrule
Texts      & 0.123   & \textbf{0.019}   & $89\%\rightarrow98\%$ \\
\hline
\end{tabular}
\caption{Log-MSE performance of S4 with HiPPO and WaLRUS cores on delay tasks. For the Texts dataset, we report cross-entropy. WaLRUS consistently outperforms HiPPO across all benchmarks.}
\label{tab:delay}
\end{table}

\subsection{Classification Tasks}

While the ablation studies demonstrate improved memory properties, we now investigate the downstream classification performance when using W4S4 initialization in full S4 models. In all cases, we keep $(A, B)$ frozen to highlight the impact of initialization.

\paragraph{Text Classification.} Given the substantial delay-task gains, we evaluate W4S4 (S4 with WaLRUS initialization) on the Texts dataset using a compact 3-layer model. Table~\ref{tab:classification_text} compares accuracy with prior works. Notably, W4S4 achieves the highest accuracy with fewer parameters and fixed $(A, B)$.

\begin{table}[ht]
\centering
\renewcommand{\arraystretch}{1.2}
\begin{tabular}{lccc}
\hline
\textbf{Model} & \textbf{Text Acc.} & \textbf{Learnable $(A,B)$} & \textbf{Params} \\
\hline
S4-LegS     & 86.82 (0.13) & \cmark & 800K \\
S4-FouT     & 86.34 (0.31) & \cmark & 800K \\
S4D-LegS    & 86.18 (0.43) & \cmark & 800K \\
S4D-Inv     & \underline{87.34 (0.20)} & \cmark & 800K \\
S4D-Lin     & 86.97 (0.23) & \cmark & 800K \\
S4 (original) & 76.02       & \cmark & 800K \\
Transformer & 64.27        & \xmark & 800K \\
\midrule
W4S4        & \textbf{88.55 (0.23)} & \xmark & \textbf{215K} \\
\hline
\end{tabular}
\caption{Text classification accuracy. W4S4 uses significantly fewer parameters and fixed $(A,B)$, yet outperforms all prior methods.}
\label{tab:classification_text}
\end{table}

\paragraph{Speech Commands Classification.}
Speech signals often span long sequences sampled at high frequencies. This makes the encoding of long-term dependencies especially important. We evaluate W4S4 on the autoregressive version of SpeechCommands and report results in Table~\ref{tab:classification_speech}.

\begin{table}[ht]
\centering
\renewcommand{\arraystretch}{1.2}
\begin{tabular}{lccc}
\hline
\textbf{Model} & \textbf{SC-ar Acc.} & \textbf{Learnable $(A,B)$} & \textbf{Params} \\
\hline
S4-LegS     & \underline{93.60 (0.13)} & \cmark & 200K \\
S4-FouT     & 91.78 (0.10)             & \cmark & 200K \\
S4D-LegS    & 93.57 (0.09)             & \cmark & 200K \\
S4D-Inv     & 93.40 (0.67)             & \cmark & 200K \\
S4D-Lin     & 93.37 (0.05)             & \cmark & 200K \\
\midrule
W4S4        & \textbf{94.37 (0.23)}    & \xmark & 260K \\
\hline
\end{tabular}
\caption{Autoregressive classification accuracy on SpeechCommands. W4S4 with frozen $(A,B)$ still achieves the highest accuracy.}
\label{tab:classification_speech}
\end{table}

\vspace{0.5em}
These classification results highlight the downstream effectiveness of W4S4-initialized SSMs, confirming that improved memory retention translates into superior task performance. They also underscore that a high-quality initialization of the SSM can outperform approaches that rely on initializing with HiPPO and subsequently training the $A$ matrix in hopes of reaching optimal performance.

\section{Conclusion and Discussion} \label{sec:conclusion}

In this work, we introduced \textbf{W4S4}, a novel State Space Model (SSM) constructed as a variant of WaLRUS from redundant wavelet frames via the SaFARi framework. WaLRUS offers a stable diagonal representation, enabling efficient computation and storage, and provides a compelling alternative to HiPPO-based SSMs in deep neural sequence models such as S4.

We began by motivating the need for better SSM initializations by analyzing the limitations of existing HiPPO variants, particularly their diagonalizability and memory efficiency. Our proposed W4S4 addresses these limitations by leveraging the representational richness of wavelet frames, while also benefiting from a significant reduction in kernel approximation cost due to its stable spectral properties.

Through controlled ablation studies, we showed that WaLRUS significantly outperforms HiPPO in memory retention tasks. Specifically, we demonstrated that even in a minimal linear setting with frozen $(A, B)$, WaLRUS achieves a \textbf{5.3$\times$} lower MSE in delay-reconstruction tasks. This advantage remains consistent across a range of delay lengths and time-invariant dynamics.

We then embedded WaLRUS into full S4 models (W4S4) and evaluated it on both synthetic delay tasks and real-world classification datasets. Notably, W4S4 outperformed HiPPO-based S4 models across all benchmarks, despite using fewer parameters and keeping $(A, B)$ fixed during training. In text and speech classification tasks, where long-range dependencies are crucial, W4S4 achieved new state-of-the-art accuracy, even compared to models with learnable $A$ matrices.

These findings underscore a key insight: \emph{high-quality initialization of the SSM core is more valuable than training $A$ with suboptimal starting points.} W4S4 provides a structured inductive bias that more accurately captures the temporal dynamics of sequential data at initialization, leading to better performance with fewer learnable parameters and faster convergence.

\paragraph{Future Work.}
Our results open several directions for future research. First, the combination of WaLRUS with learnable $(A, B)$ matrices may unlock even greater expressivity. Second, exploring other redundant frames or overcomplete bases may further enrich the SSM design space. Finally, more empirical evidence should be gradually added to this manuscript.

Overall, WaLRUS provides a principled and effective building block for long-range sequence modeling, and we believe it can serve as a foundation for the next generation of deep SSM architectures.

\bibliographystyle{unsrt} 
\bibliography{ref}
\section{Appendix}

\subsection{Theoretical findings}

\paragraph{Theorem 1.} 
Let $A$ be the diagonalized state matrix of WaLRUS constructed from a redundant wavelet frame. Then:
\begin{itemize}
    \item Only the first $N_{\mathrm{eff}}$ eigenvalues of $A$ are greater than 1; the rest are exactly 1.
    \item Only the first $N_{\mathrm{eff}}$ elements of the state vector $\widetilde{x}$ (in the diagonal basis) contribute to the reconstruction of the input signal.
    \item The kernel $K$ can be computed using only the top $N_{\mathrm{eff}}$ rows of $\widetilde{K}$ and the corresponding rows of the eigenvector matrix $V$.
\end{itemize}

This result allows us to truncate the WaLRUS state space to $N_{\mathrm{eff}}$ dimensions without sacrificing reconstruction fidelity. In practice, we:
\begin{enumerate}
    \item Diagonalize $A$ and compute only the top $N_{\mathrm{eff}}$ eigenvalues and rows of $\widetilde{B}$.
    \item Compute the kernel $\widetilde{K}$ using this reduced system.
    \item Absorb $V[0{:}N_{\mathrm{eff}}, :]$ into the output matrix $C$ and proceed with standard SSM computations.
\end{enumerate}

\subsection{Datasets} \label{appendix:datasets}

To demonstrate the improvement that W4S4 provides, we use a popular and diverse suite of datasets spanning vision, biomedical signals, speech, and multivariate time series forecasting. These datasets vary significantly in sequence length, modality, and task structure, providing a rigorous testbed for generalization and long-range dependency modeling.

\paragraph{Sequential CIFAR (sCIFAR) \cite{krizhevsky2009learning}} This task transforms the classic CIFAR-10 image classification dataset into a sequential format by flattening $32 \times 32$ RGB images into sequences of length 1024. In our experiments, we convert the RGB channels to grayscale for the simplicity. This benchmark tests a model’s ability to capture spatial structure using sequence processing.

\paragraph{BIDMC Vital Signs \cite{goldberger2000physiobank,pimentel2016toward}} Collected from the Beth Israel Deaconess Medical Center, this dataset includes physiological signals such as electrocardiogram (EKG) and photoplethysmogram (PPG), sampled over 4000 time steps. The prediction task involves estimating respiratory rate (RR), heart rate (HR), and blood oxygen saturation (SpO\textsubscript{2}). We follow prior work and focus on SpO\textsubscript{2} to make the comparison easier.

\paragraph{Speech Commands (SC) \cite{warden2018speech}.} This dataset consists of 1-second raw audio waveforms (16,000 samples) corresponding to 35 spoken words. We evaluate models in both autoregressive (AR) and Bidirectional(Bi) setting. The auroregressive setting reflects causal speech generation tasks, where state-space models have recently demonstrated strong performance.

\paragraph{Long Range Arena (LRA) \cite{tay2020long}.} LRA is a benchmark suite specifically designed to evaluate long-range dependency modeling. It includes: 
\begin{itemize} 
    \item \textbf{ListOps}: A parsing task requiring models to evaluate nested prefix expressions. 
    \item \textbf{Text (IMDB)}: Character-level sentiment classification on IMDB movie reviews. 
    \item \textbf{Retrieval}: A synthetic document matching task. 
    \item \textbf{Images}: Classification on sequences formed by flattening images from Cifar, then converting it to discrete levels from 0 to 256. \item \textbf{Pathfinder} and \textbf{PathX}: Visual reasoning tasks involving long-range spatial connectivity.
\end{itemize}

\paragraph{Multivariate Time Series Forecasting.} We also include four widely used datasets in the time series forecasting literature to evaluate long-horizon predictive performance:
\begin{itemize} 
    \item \textbf{Electricity} and \textbf{Weather}, both from \cite{wu2021autoformer}, involve hourly power consumption across 321 clients and meteorological measurements respectively. 
    \item \textbf{Traffic} and \textbf{Solar-Energy}, both from \cite{lai2018modeling}, contain occupancy rates from road sensors and solar power output across multiple stations. 
\end{itemize} 
These datasets test the model's ability to capture both seasonal trends and short-term fluctuations across multivariate signals.

\end{document}